\documentclass[journal]{IEEEtran}
\ifCLASSINFOpdf
\else
\fi
\usepackage{lineno,hyperref}
\usepackage{booktabs}
\usepackage{multirow}
\usepackage{graphicx}
\usepackage{amsfonts,bm}
\usepackage{amssymb}
\usepackage{enumerate}
\usepackage{tabularx}
\usepackage{setspace}
\usepackage{mathrsfs}
\usepackage{diagbox}

\begin{document}
%
\title{Semi-supervised mp-MRI Data Synthesis with StitchLayer and Auxiliary Distance Maximization}
%
%
%
\author{Zhiwei~Wang,~Yi~Lin,~Kwang-Ting~(Tim)~Cheng,~\IEEEmembership{Fellow,~IEEE}
	and~Xin~Yang$^*$
	\thanks{Zhiwei Wang, Yi Lin and $^*$Xin Yang are with the School of Electronic Information and Communications, Huazhong University of Science and Technology, Wuhan, 430074, China. (e-mail: xinyang2014@hust.edu.cn).}
	\thanks{K.-T. Tim Cheng is with the School of Engineering, Hong Kong University of Science and Technology, Hong Kong, China.}}

\maketitle

\begin{abstract}
In this paper, we address the problem of synthesizing multi-parameter magnetic resonance imaging (mp-MRI) data, i.e. Apparent Diffusion Coefficients (ADC) and T2-weighted (T2w), containing clinically significant (CS) prostate cancer (PCa) via semi-supervised adversarial learning. 
Specifically, our synthesizer generates mp-MRI data in a sequential manner: first generating ADC maps from 128-d latent vectors, followed by translating them to the T2w images. 
The synthesizer is trained in a semi-supervised manner. 
In the supervised training process, a limited amount of paired ADC-T2w images and the corresponding ADC encodings are provided and the synthesizer learns the paired relationship by explicitly minimizing the reconstruction losses between synthetic and real images. 
To avoid overfitting limited ADC encodings, an unlimited amount of random latent vectors and unpaired ADC-T2w Images are utilized in the unsupervised training process for learning the marginal image distributions of real images. 
To improve the robustness of synthesizing, we decompose the difficult task of generating full-size images into several simpler tasks which generate sub-images only. 
A StitchLayer is then employed to fuse sub-images together in an interlaced manner into a full-size image. 
To enforce the synthetic images to indeed contain distinguishable CS PCa lesions, we propose to also maximize an auxiliary distance of Jensen-Shannon divergence (JSD) between CS and nonCS images. 
Experimental results show that our method can effectively synthesize a large variety of mp-MRI images which contain meaningful CS PCa lesions, display a good visual quality and have the correct paired relationship. 
Compared to the state-of-the-art synthesis methods, our method achieves a significant improvement in terms of both visual and quantitative evaluation metrics.
\end{abstract}

\begin{IEEEkeywords}
Generative Models, Generative Adversarial Networks, Multimodal Image Synthesis, Deep Learning
\end{IEEEkeywords}

%
\IEEEpeerreviewmaketitle

\vspace{-0.3cm}
\section{Introduction}
\IEEEPARstart{P}{rostate} cancer (PCa) is one of the leading causes of cancer death among men~\cite{siegel2015cancer}.
In particular, men with clinically significant (CS) PCa, i.e. the Gleason Score (GS) of PCa being equal to or greater than $7$, have a much higher fatality than patients with indolent PCa.
Early detection of CS PCa is a key to increasing the survival rate of patients.
Recent studies~\cite{fehr2015automatic,lemaitre2016computer,litjens2014computer,lemaitre2015computer,wang2018automated,yang2017joint,yang2017co} have demonstrated that multi-parametric magnetic resonance imaging data (mp-MRI), which typically includes T2-weighted (T2w) images and apparent diffusion coefficient (ADC) maps could be an accurate and noninvasive biomarker for PCa detection and aggressive assessment.
However, training an accurate classifier leveraging recent advances in data-hungry deep learning, e.g. convolutional neural networks (CNNs), for PCa detection and aggressive assessment based on mp-MRI images is challenging as mp-MRI data of PCa, in particular CS PCa, are often scarce and costly to obtain.

To address this challenge, the most common approach~\cite{yang2017joint,yang2017co} is data augmentation that increases data volume by modifying the original data through rotation, translation, scaling, non-rigid transformations, etc.
However, these data augmentation approaches cannot greatly increase the data variety, limiting the performance of deep CNN models.
Recently, several medical image synthesis methods based on Generative Adversarial Networks (GAN)~\cite{salehinejad2017generalization,calimeri2017biomedical,chartsias2017multimodal,costa2017end,joyce2017robust,osokin2017gans} have been developed.
These GAN-based methods can learn the data distribution of training samples in a low-dimensional manifold, and generate new data by sampling from the learned manifold, providing an effective way to greatly increase the quantity and diversity of training data and in turn to benefit the deep learning methods.
Despite the success of existing GAN based methods, they can hardly meet the following requirements concurrently which are critical for clinical usage:
(1) ensuring correct paired relationship between synthesized ADC and T2w image pairs,
(2) increasing data variety based on a very small training set, and,
(3) containing distinguishable CS cancerous patterns in each synthesized ADC-T2w image pair.
In this work, we aim at a GAN-based method which can concurrently meet these three requirements.
In the following, we start with a survey of related work and summarize their limitations.

\textbf{Prior work:}
If we refer different modalities of MRI as different domains, mp-MRI data synthesis can then be more broadly formulated as a multi-domain or cross-domain data synthesis problem, which has been widely studied in recent years for synthesizing both natural~\cite{zhu2017toward,liu2016coupled,isola2017image} and medical~\cite{chartsias2017multimodal,costa2017end,joyce2017robust,guibas2017synthetic} images.
Existing multi-/cross-domain data synthesis methods can be categorized into three major classes:
1) \emph{cross-domain image translation} which, given a real image sampled from one domain (e.g., a T1 image), synthesizes its counterpart in another domain (e.g., a T2 image);
2) \emph{direct multi-domain image synthesis} which generates two images of different domains with a constraint on the relationship of the pair based on a common low-dimensional vector;
3) \emph{sequential multi-domain image synthesis}, which first generates images in one domain based on low-dimensional vectors, followed by \emph{cross-domain image translation} that maps them to their counterparts in another domain.

For \emph{cross-domain image translation}, Isola \emph{et al.}~\cite{isola2017image} proposed a conditional GAN (cGAN) named pix2pix which utilizes a U-Net~\cite{ronneberger2015u} as the generator to condition on an input image of one domain and generate the corresponding output image of another domain.
The L1 pixel-wise loss and a patch-based discriminator are used for ensuring both global and local consistency between real and generated images.
However, pix2pix requires a large amount of paired labels (i.e., true pairs of images) for training, which is costly, if not infeasible.
Moreover, \emph{cross-domain image translation} methods have to condition on a real input image in a reference domain and cannot concurrently synthesize new data in both domains.
Thus the diversity of the synthesized multi-domain data largely depends on the quantity and diversity of the real input data.

\begin{figure*}[htp]
	\centering
	\includegraphics[width=0.68\linewidth]{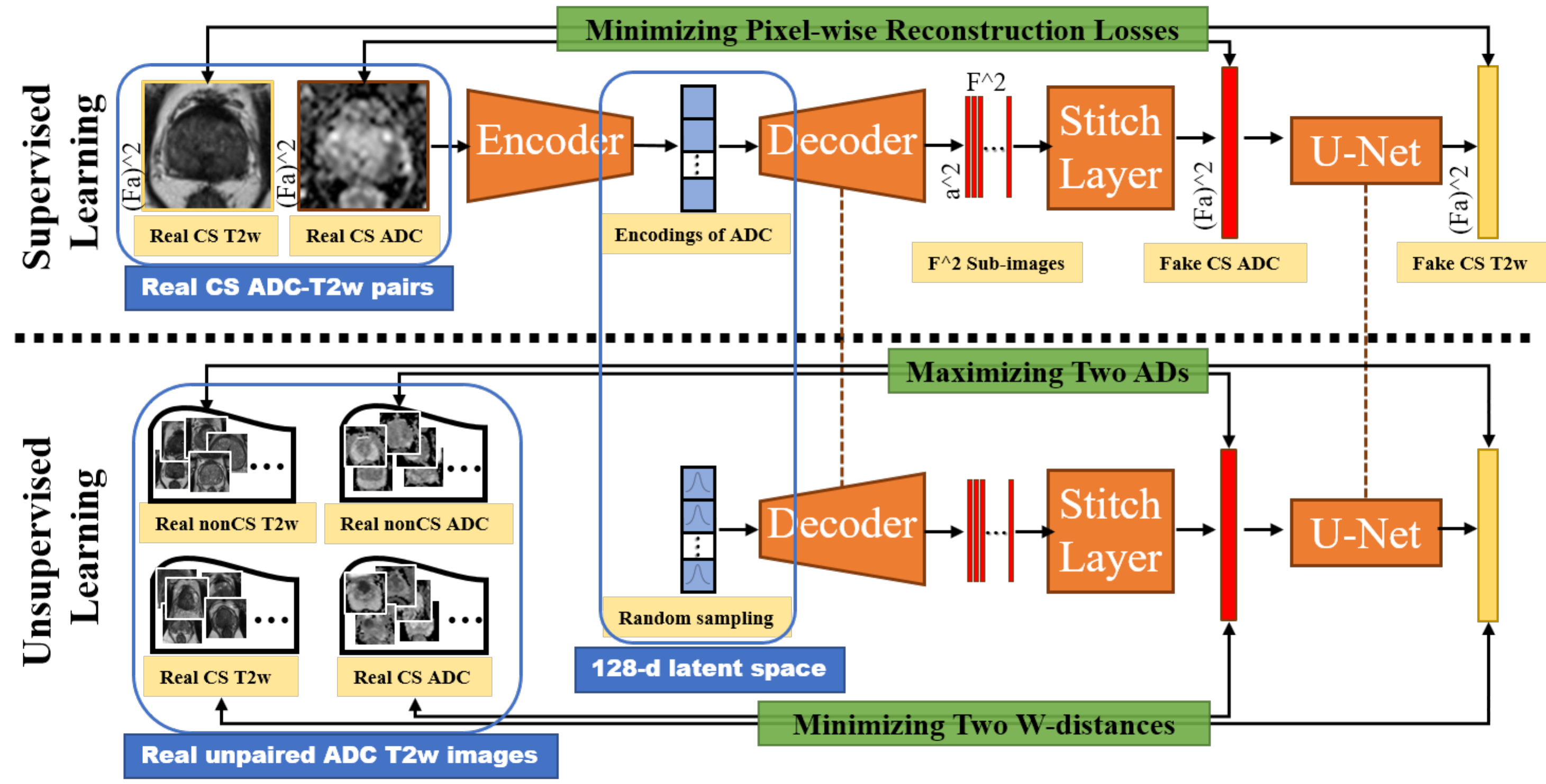}
	\caption{The framework of the proposed semi-supervised mp-MRI data synthesis method which is trained based on both supervised learning (top) for enforcing a constraint of paired relationships between synthetic ADC and T2w images, and unsupervised learning (bottom) for making our method 'see' the whole picture of the 128-d latent space so as to avoid overfitting the encodings.
		The green boxes describe the training targets of our method.
		The orange boxes connected with a dotted line share all weight parameters.}
	\label{fig:1}
\end{figure*}

Rather than relying on real input data of one domain, Liu and Tuzel~\cite{liu2016coupled} proposed a \emph{multi-domain image synthesis} method, named coupled GAN (CoGAN), for synthesizing multi-domain data directly from random noise vectors.
Specifically, they utilized two parallel identical GANs to learn the marginal image distributions in two different domains respectively.
Images of each domain are then synthesized by sampling from the corresponding marginal distribution.
To guarantee that the multi-domain images of a pair indeed follow the intended relationship, the authors proposed to share weight parameters in the parallel GANs.
Despite the success in handling their target tasks, direct mapping from random noises to multi-domain data completely ignores the inherent complexity difference between the synthesis tasks of different MRI modalities.
For instance, as shown in Fig.~\ref{fig:1}, the ADC map, capturing the functional information of a prostate, usually has a low spatial resolution and thus is easier to synthesize. 
In comparison, the T2w image which captures the anatomical structure of a prostate contains much more detailed high-frequency texture information, and thus encounters a greater challenge for the synthesis task.

To take into consideration of the differences in task complexity, Costa \emph{et al.}~\cite{costa2017end} proposed a \emph{sequential multi-domain image synthesis} method, which first tackles the simpler task, generating a retinal vessel network map, via a GAN, followed by tackling the more difficult task, synthesizing the corresponding retinal fundus image, using another GAN.
The method is trained in a supervised manner, where only the encodings of real data along with paired labels are used.
The constraint of the paired relationship between a vessel map and its corresponding retinal image is accomplished by minimizing both the reconstruction losses of vessel-fundus pairs and an adversarial loss.
However, the method in~\cite{costa2017end} can hardly capture meaningful information of CS PCa since normal prostate gland tissues typically take predominant regions compared to CS lesions, biasing the learning process to model the distribution of the predominant prostate gland rather than CS PCa.
In addition, the variety of the synthetic data is still restricted if the amount of training data is limited because the method could not generate 'unseen' data from random inputs due to its overfitting these encodings~\cite{zhu2017toward}.
To summarize, existing methods can hardly synthesize mp-MRI data with sufficient variety which contain meaningful CS PCa patterns and have the correct paired relationship from a small amount of training data.

In this paper, we propose a novel semi-supervised method which can meet all the above-mentioned requirements and synthesize high quality pairs of ADC and T2w images.
As illustrated in Fig.~\ref{fig:1}, our mp-MRI data synthesizer consists of three cascaded components: a decoder to derive low-dimensional ADC maps from 128-d latent vectors, a StitchLayer to convert low-dimensional ADC maps to a full-size ADC image, and a U-Net to convert the ADC image to a paired T2w image.
By training the synthesizer in a supervised manner which minimizes pixel-wise reconstruction losses between real and fake ADC-T2w pairs, the mapping relation between low-dimensional vectors and ADC maps, and the paired relationships can be captured by the decoder and the U-Net of the synthesizer respectively.
Training the synthesizer using supervised learning requires real pairs of CS ADC-T2w images, which could lead to overfitting a small set of ADC encodings derived from a training set of limited size~\cite{costa2017end}.
To increase the diversity of the generated data, we further train the synthesizer in an unsupervised manner through provision of various random latent vectors as illustrated in the bottom of Fig.~\ref{fig:1}.
To ensure high visual similarity between real ADC/T2w images and fake ones generated from random vectors, we minimize the Wasserstein distances (W-distance) between the marginal distributions of synthesized and real images of the two modalities.
The synthesizer is trained following supervised and unsupervised processes alternatively, i.e. in a semi-supervised fashion, to enforce paired relationships in the network.
This method can achieve a greater diversity in the generated data, and higher visual similarity between the fake and real images.

It's difficult and fragile to optimize a synthesizer, in the unsupervised training process, to directly generate full-size ADC maps from very low-dimensional latent vectors because no explicit guidance from the real ADC images would be provided.
To address this problem, we propose a StitchLayer which seamlessly fuses $F^2$ sub-regions of an ADC map into a full-size ADC image in an interlaced manner.
With the StitchLayer, the decoder could focus on tackling $F^2$ simpler tasks, each of which generates a sub-image with a smaller size (i.e., $a \times a$), instead of directly tackling the difficult task of generating a single full-size map (i.e., $Fa \times Fa$).
This strategy greatly reduces the complexity of modeling the entire manifold of ADC data.

To further ensure the synthesized data contain distinguishable CS cancerous patterns, we compute the auxiliary distances (AD) of Jensen-Shannon divergence (JSD) between the synthetic CS and real nonCS images of the two modalities.
By maximizing the two ADs, in addition to minimizing the W-distances in the unsupervised training process, the synthesizer is guided to include meaningful CS PCa features through attempting to better distinguish the synthesized CS mp-MRI data from those real nonCS mp-MRI data.

To summarize, the main contributions of this work include:

\begin{itemize}
	\item We develop a novel semi-supervised mp-MRI data synthesis method, which is trained using both supervised and unsupervised approaches alternatively.
	Supervised learning enforces the correct paired relationship between synthesized ADC and T2w images of a pair, and unsupervised learning avoids overfitting encodings and thus ensures greater diversity of the synthesized data.
	
	\item We propose a novel StitchLayer for more robustly synthesizing ADC maps in a coarse-to-fine manner.
	Consequently, the enhanced ADC map synthesis helps improve the subsequent ADC-to-T2w translation, which, in turn boosts the overall quality of synthetic mp-MRI data.
	
	\item We introduce the auxiliary distances of JSD between synthetic CS and real nonCS images for encoding clinically meaningful CS PCa-relevant visual patterns in the synthetic data.
	
	\item We conducted extensive experimental evaluations using both qualitative and quantitative metrics. 
	Experimental results demonstrate the superior performance of our method to the state-of-the-art methods and its great potential for the real clinical applications~\cite{liu2016coupled,costa2017end}.
\end{itemize}

\vspace{-0.3cm}
\section{Method}
In this section, we detail the key techniques used in our synthesizer, including 1) semi-supervised training for explicitly learning the paired relationships and modeling the marginal image distribution of each modality; 
2) a StitchLayer to alleviate the complexity of the decoder for optimizing full-size ADC map generation; and 
3) an auxiliary distance for enforcing the synthetic data containing distinguishable CS cancerous patterns.

\subsection{The Semi-supervised mp-MRI data synthesis}
Given a set of CS cancerous ADC-T2w prostate pairs $\mathbb{P}_{CS}(a,t)=\left\lbrace (a_i,t_i)|i=1,2,\dots,n\right\rbrace $ for training, where $(a_i,t_i)$ indicates the $i^{th}$ ADC-T2w pair and $n$ is the total number of pairs, a straightforward solution of adversarial training is to input random noise vectors to a synthesizer and train the synthesizer to fool the discriminator that can best distinguish synthetic mp-MRI data $(\hat{a},\hat{t})$ from real ones $\mathbb{P}_{CS}(a,t)$.
A major limitation of this solution is that the task of the discriminator could be too burdensome as it needs to not only learn the paired relationships but also ensure synthesized data being visually-realistic.
As a result, the discriminator could be difficult to train. 
An alternative solution is to input encodings of real data rather than noise vectors to the synthesizer as proposed in~\cite{costa2017end} which will then alleviate the task of the discriminator by minimizing the pixel-wise reconstruction losses between reconstructed fake ADC-T2w images and true ADC-T2w images.
However, this strategy could make the synthesizer easily overfit a small set of encodings and consequently lead to poor performance of the synthesizer for random vector inputs which are beyond the distribution of true encodings.
In this section, we present a novel semi-supervised learning to address the above-mentioned limitations.
We divide the entire task into three subtasks: 
1) learning the paired relationships in the supervised learning process via pixel-wise reconstruction loss minimization, 
2) learning the marginal distributions of the real ADC and T2w images via W-distance minimization in unsupervised training, and 
3) learning the distinguishable visual features of CS PCa via maximization of the auxiliary distance between CS and nonCS images in unsupervised training.

\subsubsection{Supervised Learning of Paired Relationships}
\begin{figure}[htbp]
	\centering
	\includegraphics[width=0.9\linewidth]{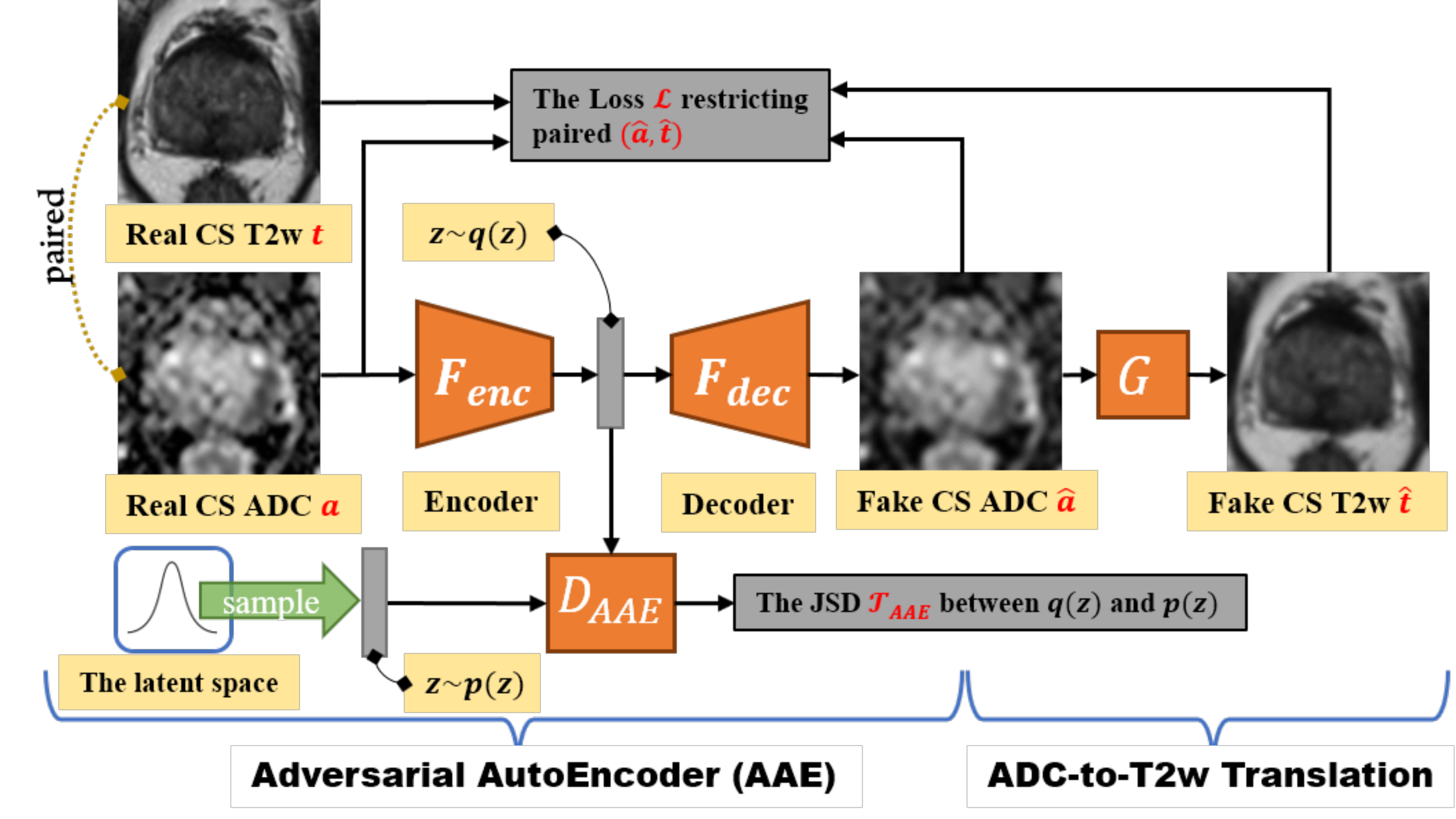}
	\caption{Supervised learning for the constraint of paired relationships between synthetic ADC and T2w image pairs.}
	\label{fig:2}
\end{figure}

Fig.~\ref{fig:2} shows details of the supervised training process.
We first utilize an encoder $F_{enc}$ to obtain encodings of real CS ADC maps, i.e. $z=F_{enc}(a)$, where $a$ is a real CS ADC map.
Then a decoder $F_{dec}$ is applied to reconstruct a fake ADC map $\hat{a}=F_{dec}(z)$ which is then converted to a fake T2w image as $\hat{t}=G(\hat{a})$.
In this work, we implement $G$ using the U-Net.
The reconstructed ADC-T2w pair $(\hat{a}, \hat{t})$ can find its target pair $(a,t)$ according to the paired labels.
We train the entire network, including the encoder $F_{enc}$ and the synthesizer (i.e. $F_{dec}$ and $G$), by minimizing the pixel-wise reconstruction loss $\mathcal{L}$ as:
\begin{equation}
\mathcal{L} = \mathbb{E}_{a,t \sim \mathbb{P}_{CS}(a,t)} \left[ || a - \hat{a} ||^{2}_{2} + || t - \hat{t} ||^{2}_{2} \right]
\label{eq:1}
\end{equation}
where $\mathbb{E}_{a,t \sim \mathbb{P}_{CS}(a,t)}$ is the expectation over the pairs $(a, t)$, sampled from the joint data distribution of real training pairs $\mathbb{P}_{CS}(a,t)$ and the operation $|| x-\hat{x}||_{2}$ calculates the average of pixel-wise Euclidean distances between images $x$ and $\hat{x}$.
As the training process is guided by real pairs of ADC-T2w images, the paired relationships can be effectively captured by the $G$ network.

To enable the synthesizer for generating reasonable ADC-T2w pairs from noise vectors rather than from ADC encodings, we adopted the approach in~\cite{makhzani2015adversarial} to reshape the distribution of encodings $q(z)$ to a pre-defined distribution $p(z)$, i.e. Gaussian distribution.
Assume that the data distribution of ADC lies on a 128-d manifold, denoted as a latent space, the reshaped distribution $q(z)$ is obtained by minimizing the JSD between the pre-defined $p(z)$ and the distribution of true ADC encodings $q(z)$.
The minimization process is implemented by designing a $D_{AAE}$ to best distinguish encodings $z \sim q(z)$ from the latent vectors $z \sim p(z)$.
Then the JSD (i.e. $\mathcal{J}_{AAE}$) between $q(z)$ and $p(z)$ is calculated as follows:
\begin{eqnarray}
\mathcal{J}_{AAE} = \max \{\mathbb{E}_{z \sim p(z)}\left[\log(D_{AAE}(z))\right] \nonumber \\
+ \mathbb{E}_{a,t \sim \mathbb{P}_{CS}(a,t)}\left[ 1-log(D_{AAE}(F_{enc}(a)))\right]\}
\label{eq:2}
\end{eqnarray}
According to~\cite{makhzani2015adversarial}, minimizing the $\mathcal{J}_{AAE}$ makes the $p(z)$ and the $q(z)$ identical to each other, which allows us to sample from the known prior $p(z)$ for synthesizing in the test phase.

Therefore, the final objective function of supervised learning is formulated as Eq.~\ref{eq:3}.
\begin{equation}
\mathcal{L}_{sup} = \mathcal{J}_{AAE} + \mathcal{L}
\label{eq:3}
\end{equation}

By minimizing $\mathcal{L}_{sup}$, our synthesizer can focus on learning the paired relationships and confining the distribution of ADC encodings to conform to a pre-defined distribution $p(z)$.

In principal, by learning the synthesizer based on the procedure described above, we could generate a variety of reasonable ADC-T2w pair from various latent vectors $z \sim p(z)$.
However, in practice the visual quality of the synthesized ADC-T2w pairs from noise vectors could be extremely poor.
We believe the reason is that the synthesizer (i.e. $F_{dec}$ and $G$) only 'sees' a very sparse and small portion of the latent space which contains the ADC encodings.
When the amount of training ADC-T2w pairs is very small, it is easy to overfit the synthesizer to the limited training samples.
To address this problem, we further apply an unsupervised learning approach to guide the synthesizer learn the marginal distributions of real ADC and T2w images and in turn to ensure a high visual quality of the synthetic ADC T2w images generated from random latent vectors.

\subsubsection{Unsupervised Learning of Marginal Distributions}
\begin{figure}[htbp]
	\centering
	\includegraphics[width=0.85\linewidth]{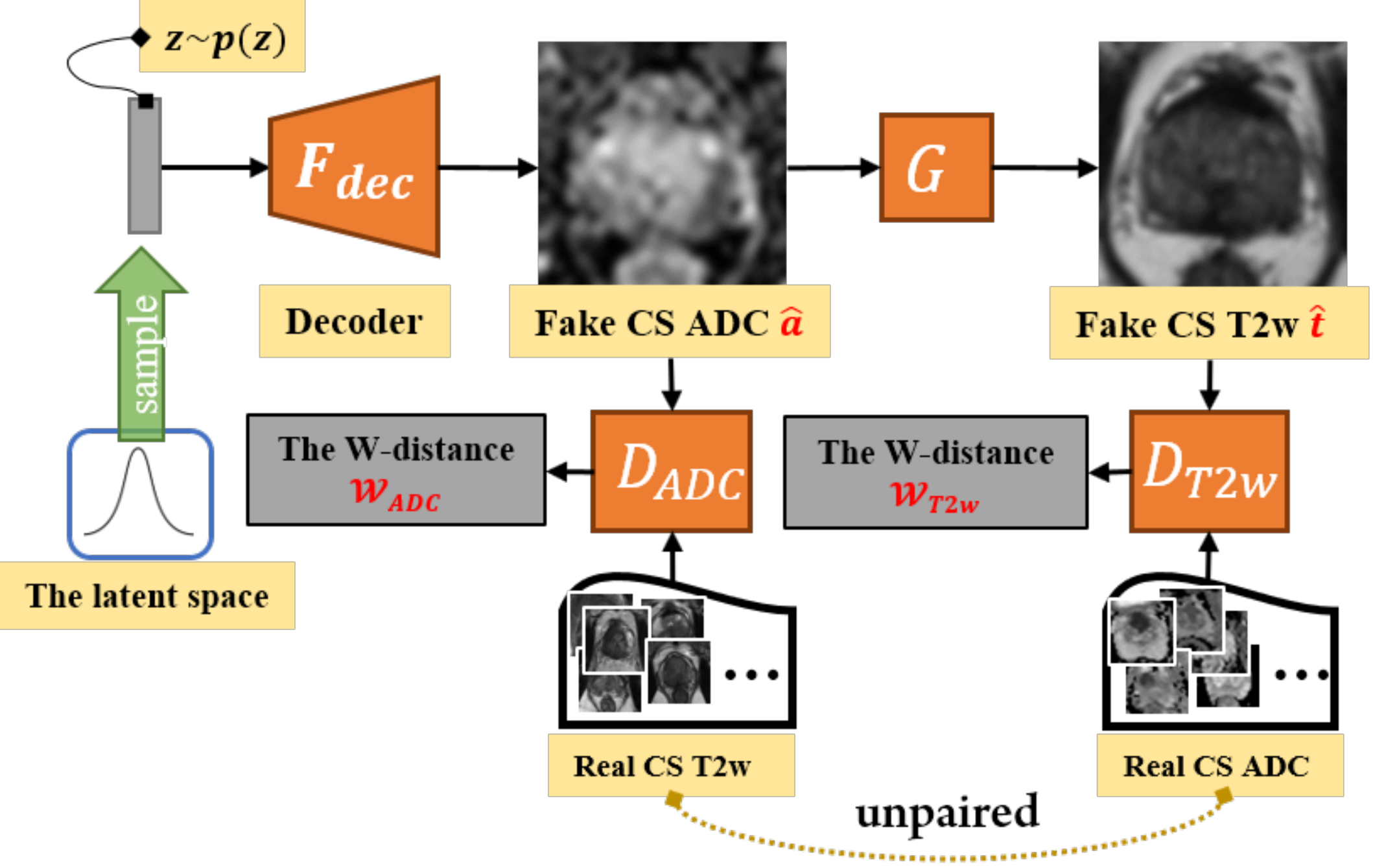}
	\caption{Unsupervised learning for mimicking the marginal distribution in each modality (i.e., ADC and T2w).}
	\label{fig:3}
\end{figure}

Fig.~\ref{fig:3} shows details of the unsupervised approach.
Compared to the supervised approach shown in Fig.~\ref{fig:2}, the unsupervised approach trains the synthesizer not based on limited encodings and paired ADC-T2w pairs, but based on unlimited latent vectors drawn from $z \sim p(z)$ and unpaired ADC images $\mathbb{P}_{CS}(a)$ and T2w images $\mathbb{P}_{CS}(t)$ of CS PCa.
We employ a discriminator $D_{ADC}$ to approximate the W-distances $\mathcal{W}_{ADC}$ between the fake and real ADC images, and another discriminator $D_{T2w}$ to approximate $\mathcal{W}_{T2w}$ as:
\begin{eqnarray}
\mathcal{W}_{ADC} = \max \{\mathbb{E}_{a \sim \mathbb{P}_{CS}(a)}\left[D_{ADC}(a)\right] \nonumber \\
- \mathbb{E}_{z \sim p(z)}\left[D_{ADC}(\hat{a})\right]-\lambda_{ADC} R_{ADC}\}
\label{eq:4}
\end{eqnarray}
\begin{eqnarray}
\mathcal{W}_{T2w} = \max \{\mathbb{E}_{t \sim \mathbb{P}_{CS}(t)}\left[D_{T2w}(t)\right] \nonumber \\
- \mathbb{E}_{z \sim p(z)}\left[D_{T2w}(\hat{t})\right]-\lambda_{T2w} R_{T2w}\}
\label{eq:5}
\end{eqnarray}
where $a$ and $\hat{a}=F_{dec}(z)$ are real and synthetic ADC maps of CS PCa respectively, $t$ and $\hat{t}=G(\hat{a})$ are real and synthetic T2w images of CS PCa respectively, $R_{ADC}$ and $R_{T2w}$ are used for enforcing the 1-Lipschitz constraint of $D_{ADC}$ and $D_{T2w}$ respectively, $\lambda_{ADC}$ and $\lambda_{T2w}$ are two parameters for adjusting the impact of $D_{ADC}$ and $D_{T2w}$ respectively~\cite{gulrajani2017improved}. 

Therefore, the final objective function of unsupervised learning is calculated as Eq.~\ref{eq:6}.
\begin{equation}
\mathcal{L}_{unsup} = \mathcal{W}_{ADC} + \mathcal{W}_{T2w}
\label{eq:6}
\end{equation}
Minimizing $\mathcal{L}_{unsup}$ can train the synthesizer to generate ADC and T2w images which conform to the marginal distributions of true ADC and T2w images respectively, i.e. visually realistic ADC and T2w images.
It is noteworthy that we do not require the unsupervised approach to learn the paired relationship between ADC and T2w of a synthetic pair. 
As such information has been captured by the supervised approach and encoded in the $G$ network. 
By alternatively training the entire network using the supervised and unsupervised approaches, our synthesizer can generate a great variety of ADC-T2w pairs that are both visually realistic and having correct paired relationships.

\vspace{-0.3cm}
\subsection{The StitchLayer for Alleviating Generation Complexity}
The quality of the synthesized ADC is critical for the following synthesis of T2w.
We observe that directly generating a full-size ADC map from a low-dimensional latent vector, i.e. 128-d vector, via a decoder is very challenging, especially for the unsupervised approach without any explicit guidance from real ADC images.
A typical matrix size of abdominal MRI scan is around $180 \times 144$, in which the prostate gland and its neighboring tissues roughly locate at the center of an ADC map and cover around $1/9$ area of the entire image. 
This implies that, to maintain the original resolution, a synthesized prostate ADC map should be at least $64\times64$.
However, most widely-used GANs~\cite{radford2015unsupervised,chen2016infogan} are limited to synthesize very small images such as images from the CIFAR-10~\cite{krizhevsky2009learning} and MNIST~\cite{lecun1998gradient} datasets whose image sizes are $32 \times 32$ and $28 \times 28$ respectively.

A potential solution to synthesize higher dimensional images is coarse-to-fine learning adopted in recent studies~\cite{wang2017high,denton2015deep,huang2016stacked}.
In these studies, customized generative networks and/or sophisticated training strategies were developed to synthesize data from low to high resolutions gradually.
However, these techniques are very time-consuming and hard to tune in their training phase, and seem overkill for synthesis of $64 \times 64$ images given that synthesis of $32 \times 32$ is just slightly beyond the capability of most plain GANs.

\begin{figure}[htbp]
	\centering
	\includegraphics[width=0.9\linewidth]{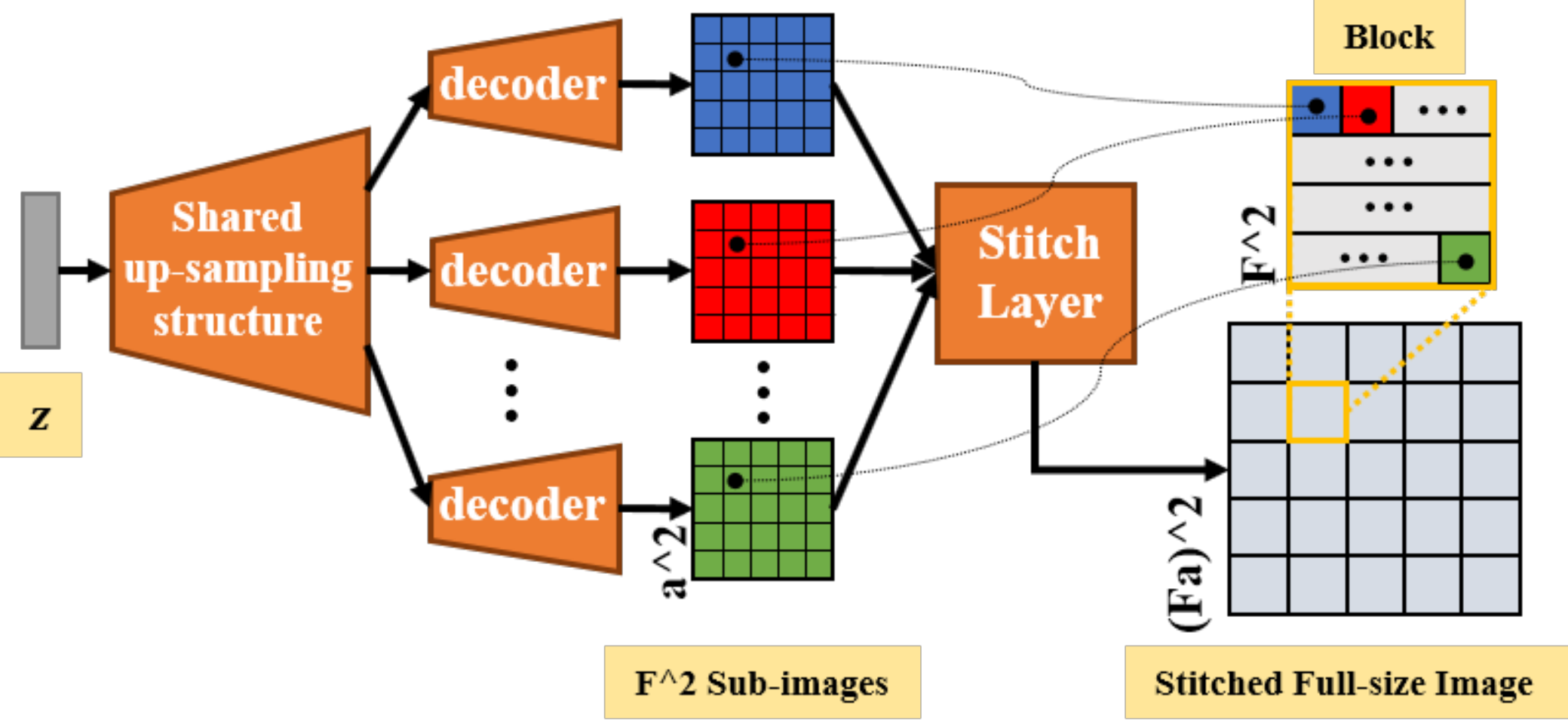}
	\caption{The StitchLayer decomposes a single hard mapping task into several simpler mapping tasks by $F^2$ decoders.
		An up-sampling structure shared among decoders learns the consistent global structure of sub-images for coarse generation of full-size image.
		Each block reconstructs local details of the full-size image from the complementary information of sub-images learned by different decoders.}
	\label{fig:4}
\end{figure}

In this work we propose a StitchLayer, which is simple yet effective, and can be embedded in any generative networks to boost existing GANs to synthesize images with a greater size.
Specifically, given the goal of generating a target image with the size of $Fa \times Fa~(F > 1)$, we start from generating smaller $a \times a$ sub-images with a total number of $F^2$ by multiple decoders as shown in Fig.~\ref{fig:4}.
Each decoder, instead of modeling a single difficult mapping task $F_{dec}(z) \rightarrow X$, where $X$ is a full-size image, only optimizes one of $F^2$ simpler mapping tasks $\{F^n_{dec}(z) \rightarrow x^n, n=1,2,\dots,F^2 \}$, where $x^n$ is the $n^{th}$ sub-image.
The StitchLayer 'stitches' these $F^2$ sub-images $\left\lbrace x^n \right\rbrace $ into a full-size ADC image $X$.
Specifically, we consider that $X$ of size $Fa \times Fa$ consists of $a^2$ non-overlapping blocks.
Each block is a square super-pixel of the size $F \times F$, which is derived as:
\begin{equation}
B_{i,j}=
\left[
\begin{array}{cccc}
x^1_{i,j} & x^2_{i,j} & \cdots & x^F_{i,j}\\
\vdots & \vdots & \vdots & \vdots \\
\cdots & \cdots & \cdots & x^{F \times F}_{i,j}\\
\end{array}
\right],i,j=1,2,\dots,a
\label{eq:7}
\end{equation}
where $B_{i,j}$ indicates a block in the $i$-th row and $j$-th column of $X$.
The block shown in Fig.~\ref{fig:4} is an exemplar of $B_{2,2}$.

In our implementation, instead of utilizing $F^2$ different decoders, all decoders share common features in the same up-sampling layers, and only differ with each other in the last fully-convolutional layer.
Accordingly, $F^2$ feature maps $\{x^n,n = 1,2,\dots,F^2\}$, i.e. sub-images of a full-size image, are generated by the decoders. 
Sharing common features in the up-sampling layers among $\{F_{dec}^n\}$ ensures globally spatial consistency among the sub-images and eliminates many unnecessary computations.
The last fully-convolutional layer of each decoder encodes unique details for each sub-image.
By 'stitching' the sub-images together, both global structure of a full-size image and complementary details of each sub-image are combined together.

Reducing the size of sub-images could make it easier to optimize the common up-sampling layers among decoders as it is easier to map a latent vector to a lower dimensional manifold.
However, reducing the sub-image's size also increases the difficulty of optimizing the decoders' last convolutional layers which capture the complementary detailed information among maps.
Therefore, it is important to choose a proper sub-image's size for a good trade-off.
In Sec.~\ref{sec:4.2} we experimentally evaluate different sub-images size for synthesizing $64 \times 64$ ADC maps.
Results show that a sub-image size of $32 \times 32$ (i.e. $F=2$) achieves the best performance.


\vspace{-0.3cm}
\subsection{Auxiliary Distance Maximization for Capturing CS PCa Patterns}
In this section we present the solution for capturing CS PCa patterns in our synthesizer which is critical for clinical usage.
The challenges of capturing CS PCa patterns in the synthesizer are twofold:
1) normal prostate gland tissues typically are predominant in a prostate image compared to a CS lesion, causing great difficulties for the synthesizer to capture sufficient CS PCa-relevant information, and
2) real CS PCa data for training is quite scarce, leading to over-fitting with a high probability.
To address these challenges, we introduce two critic networks to learn CS PCa features from a prostate gland by distinguishing between synthetic CS PCa data and real nonCS PCa data in each modality.
For both ADC and T2w, the critic networks are trained to approximate two auxiliary distances of JSD between CS PCa and nonCS PCa data respectively as follows:
\begin{eqnarray}
\mathcal{J}_{ADC} = \max \{\mathbb{E}_{a \sim \mathbb{P}_{nonCS}(a)}\left[\log (C_{ADC}(a))\right] \nonumber \\
+ \mathbb{E}_{z \sim p(z)}\left[1- \log (C_{ADC}(\hat{a}))\right]\}
\label{eq:8}
\end{eqnarray}
\begin{eqnarray}
\mathcal{J}_{T2w} = \max \{\mathbb{E}_{t \sim \mathbb{P}_{nonCS}(t)}\left[\log (C_{T2w}(t))\right] \nonumber \\
+ \mathbb{E}_{z \sim p(z)}\left[1- \log (C_{T2w}(\hat{t}))\right]\}
\label{eq:9}
\end{eqnarray}
where $C_{ADC}$ and $C_{T2w}$ are the two critic networks, $\hat{a}=F_{dec}(z)$ and $\hat{t}=G(\hat{a})$ are synthetic CS ADC and T2w images respectively, $\mathbb{P}_{nonCS}(a)$ and $\mathbb{P}_{nonCS}(t)$ are distributions of real nonCS ADC and T2w images respectively.

Once we obtain $\mathcal{J}_{ADC}$ and $\mathcal{J}_{T2w}$, we train our synthesizer to maximize both $\mathcal{J}_{ADC}$ and $\mathcal{J}_{T2w}$.
We choose JSD as the AD rather than W-distance used in the unsupervised learning is because JSD could better guide the synthesizer to increase the distance between CS and nonCS PCa data only when the synthetic data lacks CS PCa information.
This is because JSD derives no gradient unless the manifolds of synthetic CS and real nonCS PCa data align each other~\cite{arjovsky2017towards}.
Moreover, the over-fitting problem can be greatly alleviated as there are more nonCS PCa data than CS PCa data for training.
\begin{equation}
\min \left\lbrace \alpha (\mathcal{L}_{sup} + \mathcal{L}_{unsup}) \right\rbrace + \max \left\lbrace \beta (\mathcal{J}_{ADC} + \mathcal{J}_{T2w}) \right\rbrace 
\label{eq:10}
\end{equation}

The overall training target of our method shown in Fig.~\ref{fig:1} is summarized in Eq.~\ref{eq:10}, where $\alpha$ and $\beta$ are weights tuning the contributions of semi-supervised learning and the auxiliary distance maximization.
With the optimization of Eq.~\ref{eq:10}, the synthesizer is trained to generate a large variety of ADC-T2w pairs including meaningful and visually realistic CS PCa patterns besides the prostate gland, satisfying all the three requirements outlined in the introduction.

\vspace{-0.3cm}
\section{Data Preparation}

\subsection{Data Collection}
This study was approved by our local institutional review board. 
The mp-MRI data used in the study were collected from two datasets: 

\begin{enumerate}
	\item A locally collected dataset named \emph{TJPCa Dataset}~\cite{yang2017joint,yang2017co,wang2018automated} includes data conforming to the following five criteria:
	1) the data for PCa assessment were acquired between June 2014 and December 2015; 
	2) all data included either pathologically-proven PCa or benign prostatic hyperplasia (BPH) by a 12-core systematic TRUS guided plus targeted prostate biopsy which were performed within six weeks after the MRI examination; 
	3) the data were from the patients who did not receive focal therapy, hormones, or radiation prior the MRI scan; 
	4) the data include both ADC and T2w images; and 
	5) the imaging data do not include severe artifacts that made the examination non-diagnostic. 
	Indications for prostate MRI include: tumor detection for patients with clinical suspicion of prostate cancer (elevated PSA $ > $ 4.0 ng/mL and/or suspicious DRE) before biopsy, cancer staging, radiation planning, surgical planning, active surveillance, planning for biopsy targeting and evaluation of patients with a prior negative biopsy but could have continuous clinical suspicion of prostate cancer. 
	According to above criteria, we eventually collected data of $156$ patients, among whom $64$ were CS PCa patients (i.e., PCa with GS $\ge 7$) and $92$ were nonCS patients (i.e., BPH or indolent PCa).
	The mean age of CS and nonCS PCa patients are $66.6 \pm 8.5$ years old, ranging from $50$ to $88$ years old, and $69.0 \pm 8.4$ years old, ranging from $51$ to $85$ years old, respectively. 
	The median PSA value for CS and nonCS patients are $53.8$ ng/ml, ranging from $4.6$ to $1,000$ ng/ml and $11.8$ ng/ml, ranging from $0.26$ to $168.8$ ng/ml, respectively.
	
	\item A publicly available dataset named \emph{the PROSTATEx (training)}, which is the training set from the PROSTATEx challenge~\cite{prostatex,litjens2014computer,Clark2013The}, includes data of $70$ MRI-targeted biopsy-proven CS PCa and $134$ nonCS PCa patients. 
	Remaining testing data of $140$ patients from the PROSTATEx challenge are excluded from this study due to the lack of ground-truth labels (i.e. CS or nonCS). 
\end{enumerate}

In total, we have data of $360$ patients, where $226$ patients were normal, with benign prostatic hyperplasia (BPH) or indolent lesions, which are collectively referred to as nonCS PCa, and $134$ patients were with CS PCa.

All mp-MRI data in the TJPCa dataset were acquired on a 3.0 Tesla (T) whole-body unit MR imaging system (MAGNETOM Skyra, Siemens Medical Solutions, Erlangen, Germany), running software version Syngo MR D13. 
The acquisition parameters for the transverse, coronal, and sagittal T2WI TSE images were set as follows: repetition time [TR] is $6750$ ms, echo time [TE] is $104$ ms, echo train length is 16, section thickness is $3$ mm, there is no intersection gap, field of view [FOV] is $180$ mm and the image size is $384 \times 384$. The acquisition parameters for the transverse plane of DWI sequences were set as follows: $b$ values are $0$ and $1, 000$ s/mm$^2$, TR/TE are $6750$ ms/ $104$ ms, section thickness is $3$ mm, FOV is $180$ mm and the image size is $180 \times 144$. 
The ADC maps were computed from an Advanced Workstation. 
MRI protocols for data acquisition of the PROSTATEx (training) dataset are provided in~\cite{litjens2014computer}.

\vspace{-0.3cm}
\subsection{Data Preprocessing for Training and Testing}
\begin{figure}[htbp]
	\centering
	\includegraphics[width=\linewidth]{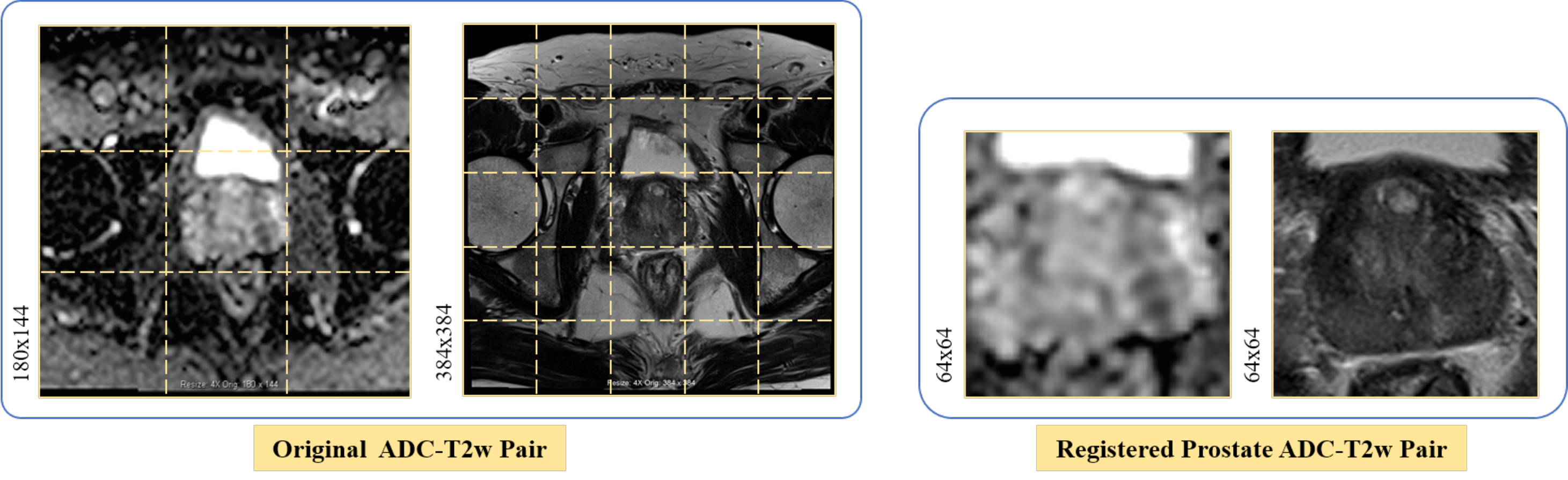}
	\caption{An example of original ADC-T2w pair (left) and its corresponding registered and cropped ADC-T2w pair of prostate region (right).}
	\label{fig:5}
\end{figure}

For these two datasets, a radiologist manually selected $533$ original ADC-T2w pairs containing CS PCa lesions and $1992$ ADC-T2w pairs which are nonCS PCa cancerous.
The selection criterion was that both CS PCa lesions and prostate glands were clearly visible.
For each selected ADC-T2w pair, we cropped and aligned the prostate region using an automated prostate detection and registration method proposed in~\cite{wang2018automated}.
As shown in Fig.~\ref{fig:5} (left), the original sizes of ADC and T2w are $180 \times 144$ and $384 \times 384$ respectively, and the width of the entire image is around $3$ and $5$ times of the width of the prostate region in ADC and T2w respectively.
Therefore, we first resized the ADC-T2w pairs to the same size of $256 \times 256$ as the input to the automated detection and registration method, and then set the output image size of the automated detection and registration method to $64 \times 64$ for better preserving useful information of the prostate in mp-MRI data.
The exemplar output of the processed prostate ADC-T2w pair is shown in Fig.~\ref{fig:5} (right).

The processed prostate ADC-T2w pairs, with a total of $2525$ pairs, were randomly divided into \emph{the TrainSet} ($483$ CS and $1942$ nonCS pairs) and \emph{the TestSet} ($50$ CS and $50$ nonCS pairs).
Each patient's data is either solely in the training or solely in the test set, but not both, to avoid overfitting data of specific patients.

\vspace{-0.3cm}
\section{Experimental Results}

\subsection{Semi-supervised Learning v.s. Supervised Learning}
We first visually compare the synthesized images produced by the semi-supervised (employing both top and bottom parts of Fig.~\ref{fig:1}) and supervised (employing only top part of Fig.~\ref{fig:1}) methods to qualitatively demonstrate the effectiveness of our proposed semi-supervised method for addressing the overfitting problem.
To better focus this analysis on only the two learning approaches, we excluded the StitchLayer and auxiliary distance maximization from the network shown in Fig.~\ref{fig:1} for both learning approaches.
To further improve the performance of the supervised method, we adopt a discriminator used in~\cite{costa2017end} which distinguishes fake pairs from real pairs for reducing the blurs in synthetic images.
Both semi-supervised and supervised methods are trained based on $483$ CS ADC-T2w pairs from the TrainSet, and latent vectors used for synthesis were obtained by two different approaches, i.e., spherical interpolation~\cite{costa2017end} and random sampling.

\begin{figure}[htbp]
	\centering
	\includegraphics[width=\linewidth]{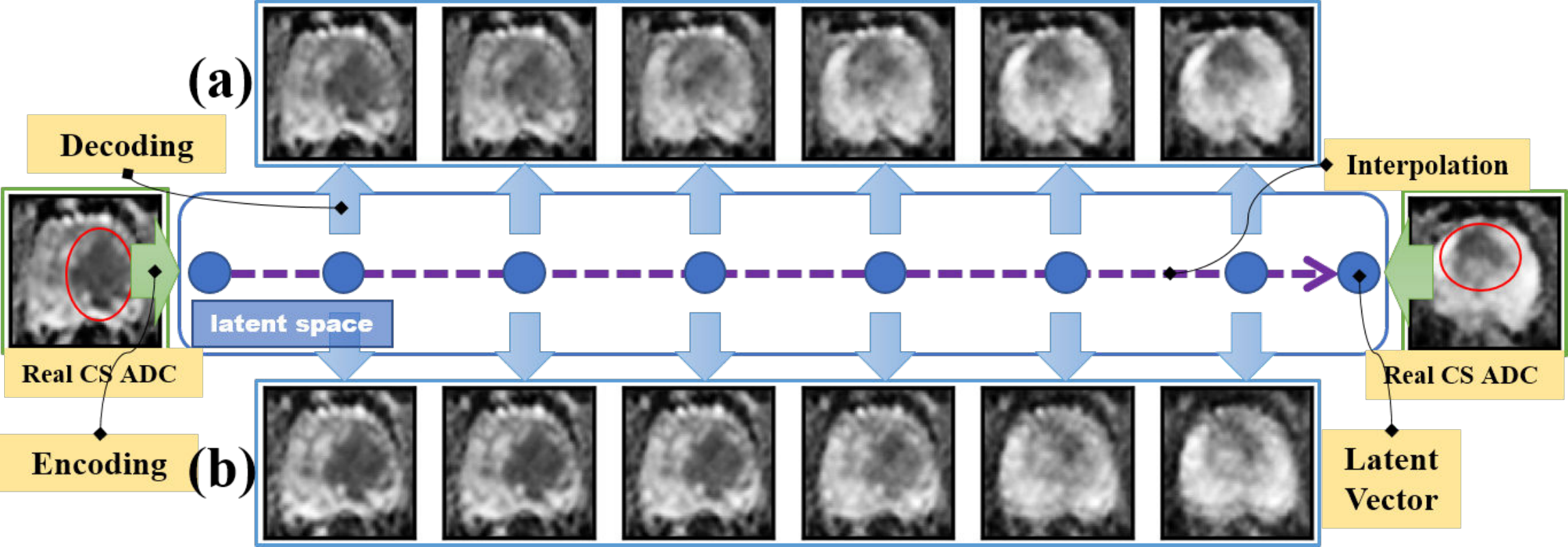}
	\caption{The synthesis methods generate ADC maps based on latent vectors obtained by the spherical interpolation.
		The ADC maps in (a) were synthesized by the semi-supervised method and the maps in (b) were synthesized by the supervised method.}
	\label{fig:6}
\end{figure}

The spherical interpolation approach is shown in Fig.~\ref{fig:6}.
The left most and right most blue dots denote ADC encodings in the latent space derived from two real CS ADC maps.
By interpolating additional dots between the two encodings, we could generate a set of new latent vectors (i.e. $2^{nd} - 7^{th}$ blue dots), based on which new fake ADC images can be generated via the decoder.
A decoder learns a complete mapping relationship between latent vectors and ADC images should be able to generate smoothly transitional images from interpolated vectors between every two real images.
To validate this, we purposely select two real ADC maps (i.e. the leftmost and rightmost images of Fig.~\ref{fig:6}) from the TestSet with a single CS PCa lesion locating on the right (in the leftmost image) and the top (in the rightmost image) of the prostate gland respectively.
The lesions are visually darker than surrounding tissues as denoted by the red circles.
Figs.~\ref{fig:6}(a) and (b) show synthesized ADC maps based on interpolations by the semi-supervised and supervised synthesizers respectively.
As seen in Fig.~\ref{fig:6}(a), the CS PCa lesion is gradually and smoothly transitioned from the right to the top in the prostate gland, while the first three images of CS PCa in Fig.~\ref{fig:6}(b) are almost identical to the leftmost real image and the transition from the $4^{th}$ image (i.e. lesion on the right) to the $5^{th}$ image (i.e. lesion at the top)  is sudden and not smooth.

\begin{figure}[htbp]
	\centering
	\includegraphics[width=\linewidth]{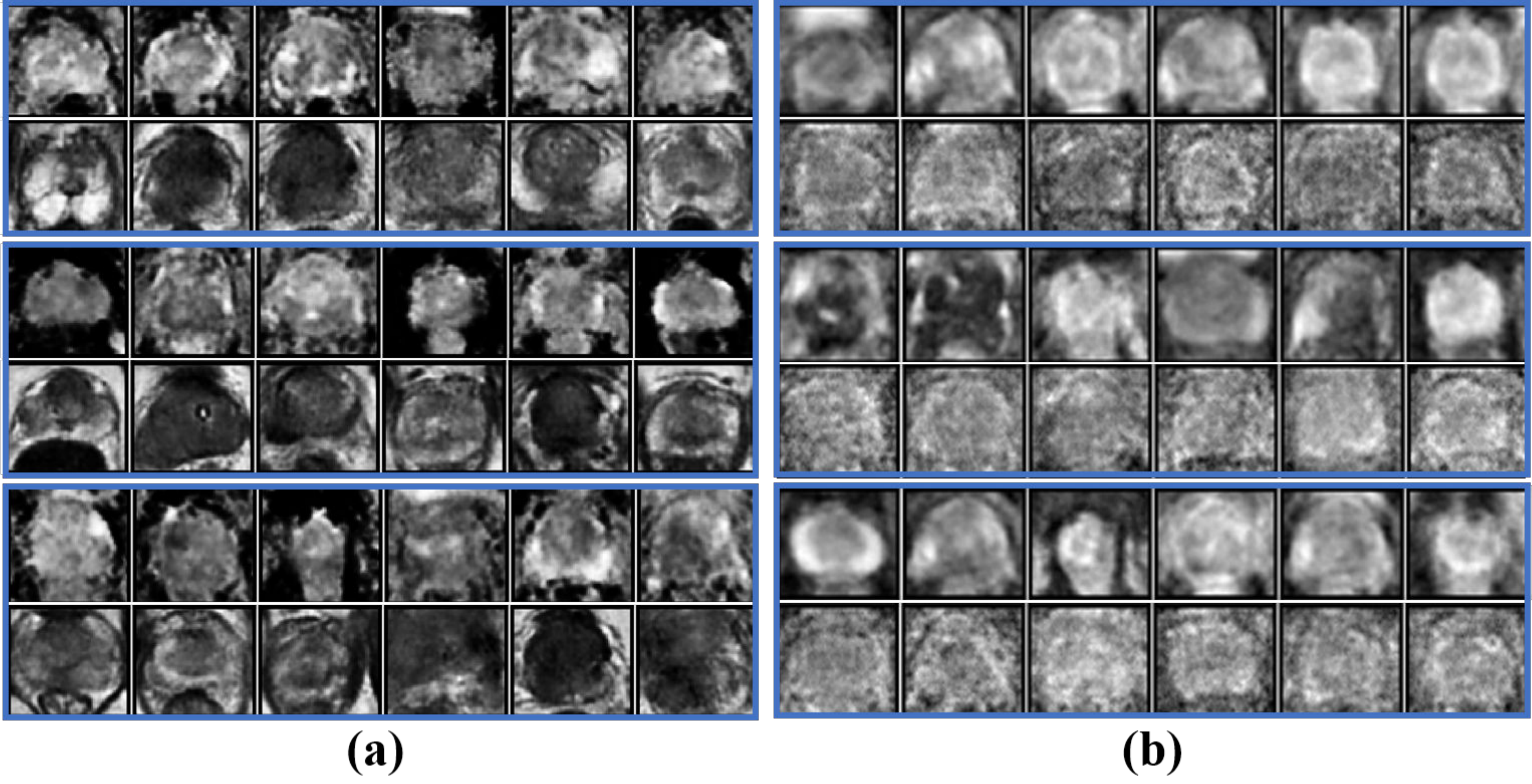}
	\caption{The synthesis methods generate ADC-T2w pairs based on a same set of random latent vectors.
		The ADC-T2w pairs in (a) were synthesized by the semi-supervised method and the pairs in (b) were synthesized by the supervised method.}
	\label{fig:7}
\end{figure}

We also utilize latent vectors randomly sampled from the prior Gaussian distribution $p(z)$ for synthesis.
Figs.~\ref{fig:7}(a) and (b) show synthetic pairs generated by the semi-supervised and supervised methods respectively.
The top row in each blue box indicates synthetic ADC maps and the bottom row indicates their corresponding generated T2w images.
As can be seen, the shapes of the prostate glands generated by the semi-supervised method are much clearer and have greater variety than those from the supervised method. 
There exists a severe mode collapse in the synthesized pairs by the supervised method especially for T2w images.
By comparing the results in Figs.~\ref{fig:6}(b) and Fig.~\ref{fig:7}(b), we notice that the supervised method can only produce somewhat realistic results for latent vectors around the encodings, but the performance degrades significantly for randomly sampled vectors in the latent space, which could greatly limit the variety of the synthesized data.
In comparison, the semi-supervised method which further guides the synthesizer to learn more complete distributions of real ADC and T2w images could facilitate the synthesizer to generate a large variety of visually-realistic data in both modalities.

\vspace{-0.25cm}
\subsection{Up-sampling Architecture and Parameter Setting of the StitchLayer}
\label{sec:4.2}
\begin{figure}[htbp]
	\centering
	\includegraphics[width=\linewidth]{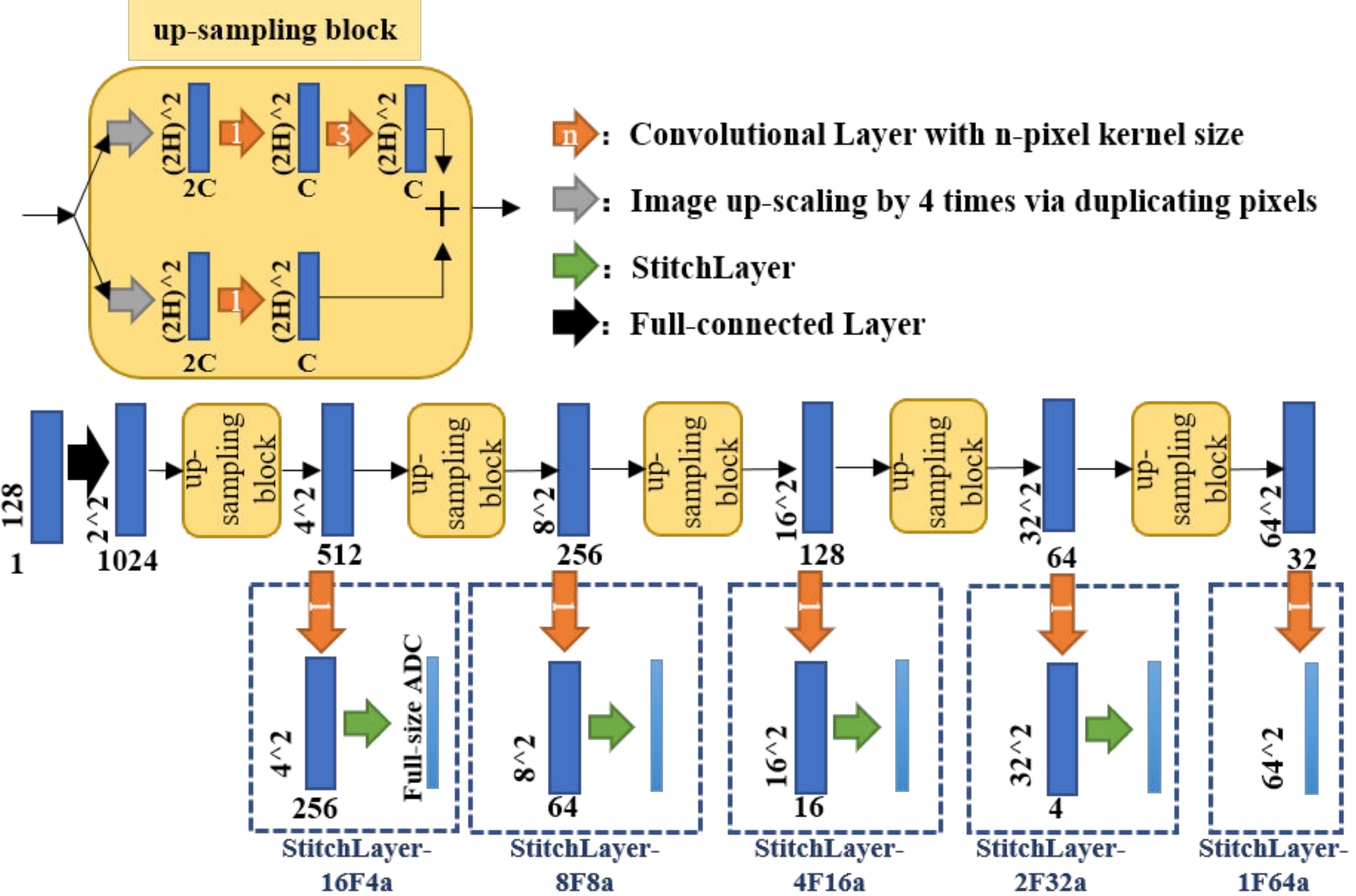}
	\caption{A specially designed up-sampling architecture is utilized for the exploration of parameter setting of the StitchLayer. Note that the StitchLayer-1F64a directly generates the full-size ADC maps without using the StitchLayer.}
	\label{fig:8}
\end{figure}

In this section, we explore the optimal setting for the StitchLayer, i.e. the size of sub-images (i.e., $a$) and number of blocks (i.e., $F$).
The up-sampling architecture of the decoder is shown in Fig~\ref{fig:8}.
Specifically, we first extract the intermediate feature maps with a size of $a \times a$ from the architecture, and then utilize a full-convolutional layer with $F^2$ kernels as decoders to obtain sub-images, followed by the StitchLayer to 'stitch' them into a full-size ADC map.
Based on the up-sampling architecture, we built different StitchLayer-based models with different parameter settings, denoted as StitchLayer-\#F\#a.
For example, as shown in Fig.~\ref{fig:8}, the StitchLayer-2F32a utilizes 4 kernels to fully convolve the penultimate feature maps with the size of $32 \times 32$, yielding 4 sub-images with the same size, and then uses the StitchLayer to obtain the full-size ADC map. 
For a fair comparison, those five StitchLayer-based models share the first fully-connected layer and up-sampling block, which together account for 77\% parameters of the up-sampling architecture, and thus have almost identical model complexity and learning ability.

\begin{figure}[htbp]
	\centering
	\includegraphics[width=\linewidth]{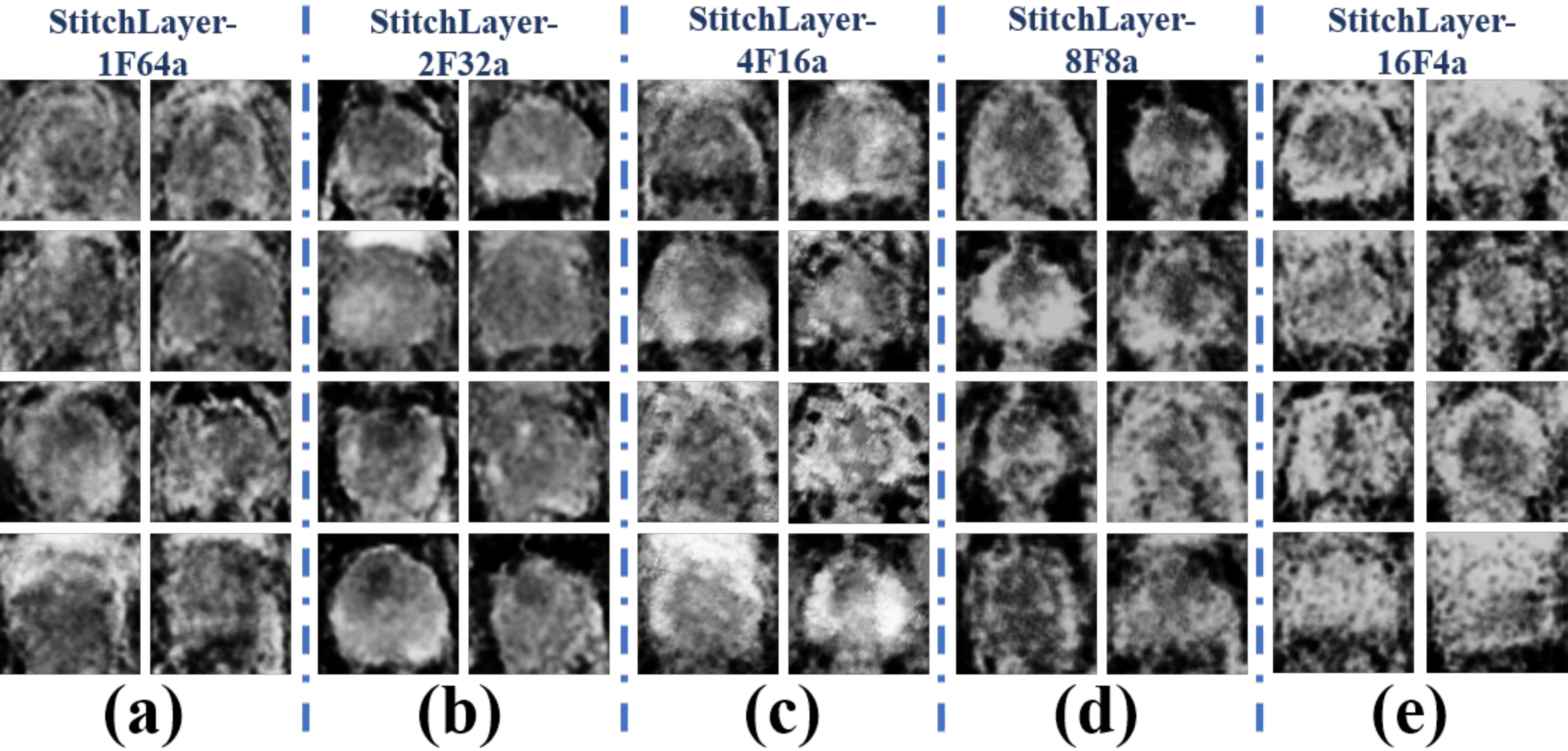}
	\caption{Examples of synthesized ADC maps by the different StitchLayer-based models.}
	\label{fig:9}
\end{figure}

We trained five StitchLayer-based models based on ADC maps from the TrainSet in an unsupervised manner, and then used them to synthesize ADC maps based on the same set of random latent vectors.
Fig.~\ref{fig:9} shows synthetic ADC maps from StitchLayer-based models with different selections of $(F,a)$.
From the outputs of the StitchLayer-1F64a, shown in Fig.~\ref{fig:9}(a), we observe a slight mode collapse problem as the first four maps are almost identical and the shapes of some prostate glands are quite ambiguous.
The StitchLayer-1F64a actually optimizes a direct generation using a single decoder.
Therefore, the decoder has to learn both global structure and local details of the full-size image, making the generation hard and fragile for optimization.

From Figs.~\ref{fig:9}(b)-(e), we observe that, for $F > 2$, a smaller $F$ results in visually more realistic and satisfactory ADC maps.
Increasing $F$ to greater than $8$ could yield comparable or even worse results than Fig.~\ref{fig:9}(a).
Our conjecture is that a larger $F$ increases the difficulties for reconstructing complementary local details.
In addition, a large $F$ also reduces the amount of global structure information preserved in the sub-images, and in turn yields large noises in both global shapes of prostate glands and local tissue patterns.

\begin{table}[htbp]
	\centering\caption{Quantitative evaluation results of the different StitchLayer-based models.}
	\begin{footnotesize}
		\begin{tabularx}{\linewidth}{*5{>{\centering\arraybackslash}X}}
			\toprule
			\specialrule{0em}{1pt}{1pt}
			StitchLayer-1F64a & StitchLayer-2F32a & StitchLayer-4F16a & StitchLayer-8F8a & StitchLayer-16F4a  \\
			\specialrule{0em}{1pt}{1pt}
			\hline
			\specialrule{0em}{1pt}{1pt}
			87\% & \textbf{90\%}  & 89\% & 86\% & 84\%  \\
			\specialrule{0em}{1pt}{1pt}
			\bottomrule
		\end{tabularx}
	\end{footnotesize}
	\label{tab:1}
\end{table}

We further quantitatively evaluate the performances of the five StitchLayer-based models in a specific task of slice-level CS vs. nonCS PCa classification.
For each model, we combine the $1942$ synthesized ADC maps of CS PCa and $1942$ real ADC maps of nonCS PCa from the TrainSet to train an Artificial Neural Network (ANN), which consists of two fully-connected layers, as the classifier.
We test the trained classifiers on the TestSet and use the classification accuracy as the metric.
The more realistic synthetic ADC maps of CS PCa are used for training, the higher classification accuracy on the TestSet can be achieved.
The classification results in Table.~\ref{tab:1} are consistent with the visual qualities of synthetic ADC maps shown in Fig.~\ref{fig:9}.
Therefore, based on both visual and quantitative evaluation results, we adopt the StitchLayer-2F32 as our generation model of ADC maps in the following experiments.


\vspace{-0.3cm}
\subsection{Comparison with the State-of-the-arts}
We compare our semi-supervised synthesis method of mp-MRI data with two state-of-the-art methods~\cite{liu2016coupled,costa2017end}.
The CoGAN proposed in~\cite{liu2016coupled} in an unsupervised method for data synthesis in multi-domain, and the method proposed by Costa \emph{et al.}~\cite{costa2017end} is trained in a supervised manner.
To the best of our knowledge, our work is the first proposal that adopts the semi-supervised approach for mp-MRI data synthesis.
Besides the two state-of-the-arts, we also compare two variants of our method which are trained with and without the auxiliary distance (AD) maximization respectively to evaluate the effectiveness of AD maximization.
The synthesized mp-MRI data by different methods are evaluated to test and verify the following three characteristics:
i) paired relationship,
ii) variety, and
iii) distinguishability of CS PCa.
Accordingly, we respectively chose the Fr\'{e}chet Inception distance (FID)~\cite{heusel2017gans}, the inception score (IS)~\cite{salimans2016improved}, and the slice-level classification accuracy (SCA) as the evaluation metrics.
For each synthesis model, we randomly generated $5$ sets of mp-MRI data of CS PCa, each of which contains $1942$ synthetic ADC-T2w pairs, and reported the average value (Avg) and the standard deviation (Std) of synthetic datasets for each metric.
Furthermore, statistical significance testing based on the t-test was performed to evaluate the statistical significance when making the comparison.

FID is a widely used metric for evaluating the distance between the distributions of the synthetic data and real data. 
Specifically, FID calculates the Wasserstein-2 distance between the generated pairs and the real pairs in the feature space of an Inception-v3 network~\cite{szegedy2016rethinking}.
Synthetic ADC maps and T2w images which are more realistic and have stronger paired relationships should have a higher probability of coming from a joint data distribution similar to the real joint data distribution of multimodalties, yielding lower FID values.

The IS is an alternative to human annotators which can automatically measure the visual quality and diversity of synthesized samples~\cite{salimans2016improved}.
The IS first uses an Inception-V3 model pre-trained on the ImageNet dataset to assign each synthetic image a class label, and then calculate the KL divergence between the conditional class distribution and the marginal class distribution.
Higher IS values indicate better quality and diversity of synthetic data.
Although the pre-trained Inception-V3 model cannot produce matching labels for our PCa data due to a different dataset for training, it is still meaningful and insightful to use IS since some common features are shared among both medical and natural data (e.g., edges, blob patterns, brightness, etc.), and more varied PCa data could be assigned more different labels by the Inception-V3 model, yielding higher IS values.
We calculated two IS values, denoted as IS-ADC and IS-T2w respectively, to measure the data variety in two modalities separately.

Inspired by the previous studies of~\cite{zhang2016colorful,costa2017end}, we introduce a task-specific evaluation metric to verify whether the model captures CS PCa-relevant information during synthesis.
Specifically, we used $1942$ synthetic ADC-T2w pairs of CS PCa and $1942$ real ADC-T2w pairs of nonCS PCa from the TrainSet to train a multimodal ANN-based classifier, which takes the ADC-T2w pair as input and predicts its a probability of being CS cancerous.
The SCA on the TestSet is then used as the metric to evaluate distinguishability of CS PCa of the synthetic data.
Higher SCA values imply that the corresponding method can synthesize ADC-T2w pairs with more distinguishable CS PCa patterns and in turn lead to a more accurate multimodal classifier.

\begin{table*}[htbp]
	\centering\caption{Quantitative comparison results (Avg $\pm$ Std) among the synthetic mp-MRI data derived from two variants of our mp-MRI synthesis method and two state-of-the-art methods, and the real data (i.e. augmented TestSet).}
	\begin{footnotesize}
	\begin{tabularx}{0.9\linewidth}{p{5cm}*4{>{\centering\arraybackslash}X}}
		\toprule
		\specialrule{0em}{1pt}{1pt}
		mp-MRI data Synthsis Method&IS-ADC&IS-T2w & FID & SCA \%   \\
		\specialrule{0em}{1pt}{1pt}
		\hline
		\specialrule{0em}{1pt}{1pt}
		CoGAN~\cite{liu2016coupled} &$1.90 \pm 0.05$&$1.77 \pm 0.01$&$231.2 \pm 7.5$&$89.0 \pm 0.3$ \\
		\specialrule{0em}{1pt}{1pt}
		Costa \emph{et al.}~\cite{costa2017end} &$1.53 \pm 0.02$&$1.63 \pm 0.02$&$239.8 \pm 6.8$&$71.6 \pm 0.5$  \\
		\specialrule{0em}{1pt}{1pt}
		\hline
		\specialrule{0em}{1pt}{1pt}
		Ours w/o the AD Maximization &$2.00 \pm 0.04$&$1.91 \pm 0.07$&$182.4 \pm 6.5$&$90.6 \pm 0.9$  \\
		\specialrule{0em}{1pt}{1pt}
		Ours w/ the AD Maximization &\textbf{2.24} $\pm$ \textbf{0.03}&\textbf{2.10} $\pm$ \textbf{0.05}&\textbf{178.2} $\pm$ \textbf{3.7}&\textbf{94.4} $\pm$ \textbf{0.5} \\
		\specialrule{0em}{1pt}{1pt}
		\hline
		\specialrule{0em}{1pt}{1pt}
		Real Data &3.27&3.26&143.8 &93  \\
		\specialrule{0em}{1pt}{1pt}
		\bottomrule
	\end{tabularx}
	\end{footnotesize}
	\label{tab:2}
\end{table*}

Table~\ref{tab:2} shows the comparison results of different mp-MRI data synthesis methods from which four observations can be made:
\begin{enumerate}[i)]
	\item By comparing the 1$^{st}$ and 2$^{nd}$ rows, we observe that the unsupervised method~\cite{liu2016coupled} significantly outperforms the supervised method~\cite{costa2017end} ($p < 0.0001$ for IS-ADC, IS-T2w and SCA, and $p = 0.0206$ for FID) since the overfitting problem prevents the synthesizer from generating realistic data with sufficient variety for random latent vectors.
	
	\item By comparing the 1$^{st}$ and 3$^{th}$ rows, we observe that our method achieves much lower FID ($p < 0.0005$) value than the CoGAN while the IS values are comparable ($p < 0.05$ for both IS-ADC and IS-T2w), implying that sharing weights is too weak to restrict paired relationships compared to minimizing pixel-wise reconstruction losses.
	Based on these first two observations, we can conclude that our semi-supervised method can combine the strengths of the unsupervised and supervised methods, and thus produce more realistic, varied and paired mp-MRI data, which meets the first two requirements of data characteristics outlined in Introduction for clinical usage.
	
	\item Comparing the two variants of our method shown in the 3$^{rd}$ and 4$^{th}$ rows, we observe that ours with the AD maximization achieves a higher SCA value than that without the AD maximization ($p < 0.0001$).
	The results confirm that the proposed AD maximization indeed helps our method learn to generate ADC-T2w data with more distinguishable CS PCa patterns, which meets the last required data characteristic outlined in Introduction.
	
	\item We further evaluated the TrainSet with respect to these metrics which are presented in the $5^{th}$ row.
	For fairness, we augmented real CS PCa data to $1942$ using the data augmentation approach proposed in~\cite{yang2017co} for training the multimodal classifier.
	By comparing all rows, we observe that the Real Data achieves the highest IS values and the lowest FID value among all synthesis methods, implying that there still exists room for improvement in synthesizing truly realistic and varied mp-MRI data.
	However, the comparison results of SCA are encouraging.
	The classifier trained with the synthetic data from ``Ours w/ the AD Maximization'' achieves a slightly better performance than that with real ones, implying that our method could synthesize data with in-depth features and is a more viable alternative for addressing the insufficiency of medical data than the traditional data augmentation for specific clinical tasks.
\end{enumerate}

\vspace{-0.3cm}
\section{Conclusions and Future Work}
Despite of a large amount of GAN-based image synthesis methods in the literature of computer vision, few of them can be directly applied to multimodal medical image synthesis tasks which possess unique challenges. 
In this study, we take the task of generating mp-MRI data of CS PCa as an application driver, to carefully study the limitations of existing methods and propose a list of novel techniques for generating clinically meaningful ADC-T2w images under the constraint of limited amount of training data. 
First, we propose a semi-supervised method to enable the synthesizer to comprehensively understand the entire latent space consisting of random vectors and encodings, and thus learn to generate an unlimited number of varied and paired ADC-T2w images based on a limited amount of real data for training. 
Second, we propose the StitchLayer which can be easily integrated into any synthesizer for alleviating the complexity of direct mapping from a low-dimensional noise to a full-size image without explicit supervision from ground-truth images. 
Third, to encode distinguishable CS cancerous visual patterns in the synthetic mp-MRI data, we propose to maximize an auxiliary distance between the real nonCS and the synthetic CS images in each modality, which enforces the synthesis process to increase the reliance on the clinically meaningful CS PCa-relevant features rather than the dominant prostate or bladder tissues. 
We collected pathology proven mp-MRI data from both a local hospital and public datasets. Visual and quantitative experimental results demonstrate that our synthesizer achieves superior performance to the state-of-the-art methods~\cite{liu2016coupled,costa2017end} and can generate ADC-T2w pairs with a great variety, with correct paired relationships and containing distinguishable CS cancerous patterns. 
Even more encouraging, our synthetic ADC-T2w data can help boost the performance of a specific clinical task (i.e. slice-level CS vs. nonCS classification) compared to relying only on the real data. 
In our future work, we could investigate the performance of synthesizing three or even more modalities in mp-MRI, e.g. ADC, T2w and Dynamic Contrast Enhanced MRI (DCE-MRI).
Our future work also includes extending the 2D synthesizer to 3D to better capture more comprehensive 3D information of mp-MRI data. 
In addition, the proposed techniques should be also applicable to the task of synthesizing many other types of medical imaging data.
Our future work will explore the potentials of our method in more medical imaging applications.

\vspace{-0.3cm}
\bibliographystyle{IEEEtran}
\bibliography{mybibfile}

\end{document}